\documentclass[letterpaper, 10 pt, conference]{ieeeconf}

\IEEEoverridecommandlockouts                              

\let\labelindent\relax
\usepackage{enumitem}
\usepackage{balance}
\usepackage[font={small}]{caption}
\usepackage{subcaption}
\usepackage{array}
\usepackage{textcomp}
\usepackage{mathtools, nccmath}
\usepackage{graphicx}
\usepackage{amsfonts}
\usepackage{amsmath}
\usepackage{amssymb}
\usepackage{algorithm}
\usepackage{algorithmic}
\usepackage{hyperref}
\usepackage{tikz}
\usepackage{arydshln}
\usepackage{multirow}
\usepackage{bm}
\usepackage{epstopdf}
\usepackage{cite}
\usepackage{pifont}
\usetikzlibrary{positioning}

\title{\LARGE \bf
Animated Cassie: A Dynamic Relatable Robotic Character
}

\author{Zhongyu Li, Christine Cummings and Koushil Sreenath
\thanks{All the authors are with the Department of Mechanical Engineering, University of California, Berkeley, CA, USA.  
        {\tt\footnotesize \{zhongyu\_li, christine.cummings, koushils\}@berkeley.edu}}%
}

\begin{document}

\maketitle
\thispagestyle{empty}
\pagestyle{empty}

\begin{abstract}

Creating robots with emotional personalities will transform the usability of robots in the real-world. As previous emotive social robots are mostly based on statically stable robots whose mobility is limited, this paper develops an animation to real-world pipeline that enables dynamic bipedal robots that can twist, wiggle, and walk to behave with emotions. First, an animation method is introduced to design emotive motions for the virtual robot's character. Second, a dynamics optimizer is used to convert the animated motion to dynamically feasible motion. Third, real-time standing and walking controllers and an automaton are developed to bring the virtual character to life. This framework is deployed on a bipedal robot Cassie and validated in experiments. To the best of our knowledge, this paper is one of the first to present an animatronic dynamic legged robot that is able to perform motions with desired emotional attributes. We term robots that use dynamic motions to convey emotions as \textit{Dynamic Relatable Robotic Characters}.

\end{abstract}


\section{Introduction and Related Work}
Pixar characters are full of life and are emotive. However, they are virtual characters and only live in our screens.  Robots living in real world are inanimate characters designed to autonomously execute tasks.  For robots to meaningfully interact with humans, they need to go beyond task execution and incite emotion in humans through their actions.  Robots need to be \textit{relatable}.  Most state-of-the-art social robots are statically stable~\cite{venture2019robot}, \textit{i.e.}, they are either small mobile robots~\cite{AnkiWeb}, or person-size ones with heavy bases~\cite{pandey2018mass,ishiguro2001robovie}, or lack mobility~\cite{PARO2006,kkedzierski2013emys}. Their capacity to convey emotion is limited to simplistic visual cues, like facial expressions, and cannot use body language to interact with humans. Conversely, dynamically stable robots, specifically bipedal robots, can move and convey emotion with much more agility and success. Unfortunately, as legged robots are high-dimensional, nonlinear and underactuated systems with hybrid dynamics, this kind of human-robot interaction has not been explored.

\begin{figure}[!]
\centering
\hspace*{0.1em}
\begin{subfigure}{.4\linewidth} 
  \includegraphics[width=1.0\linewidth,height=0.83\linewidth]{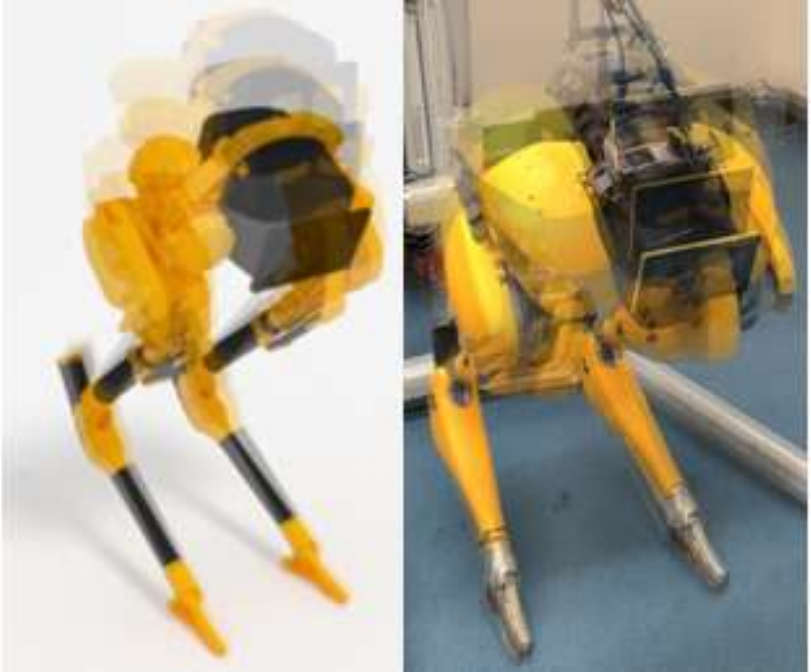}
  \caption{Tired: Nodding}
  \label{fig:sfig1}
\end{subfigure}%
\hspace*{0.1em}
\begin{subfigure}{.4\linewidth}
  \includegraphics[width=1.0\linewidth,height=0.83\linewidth]{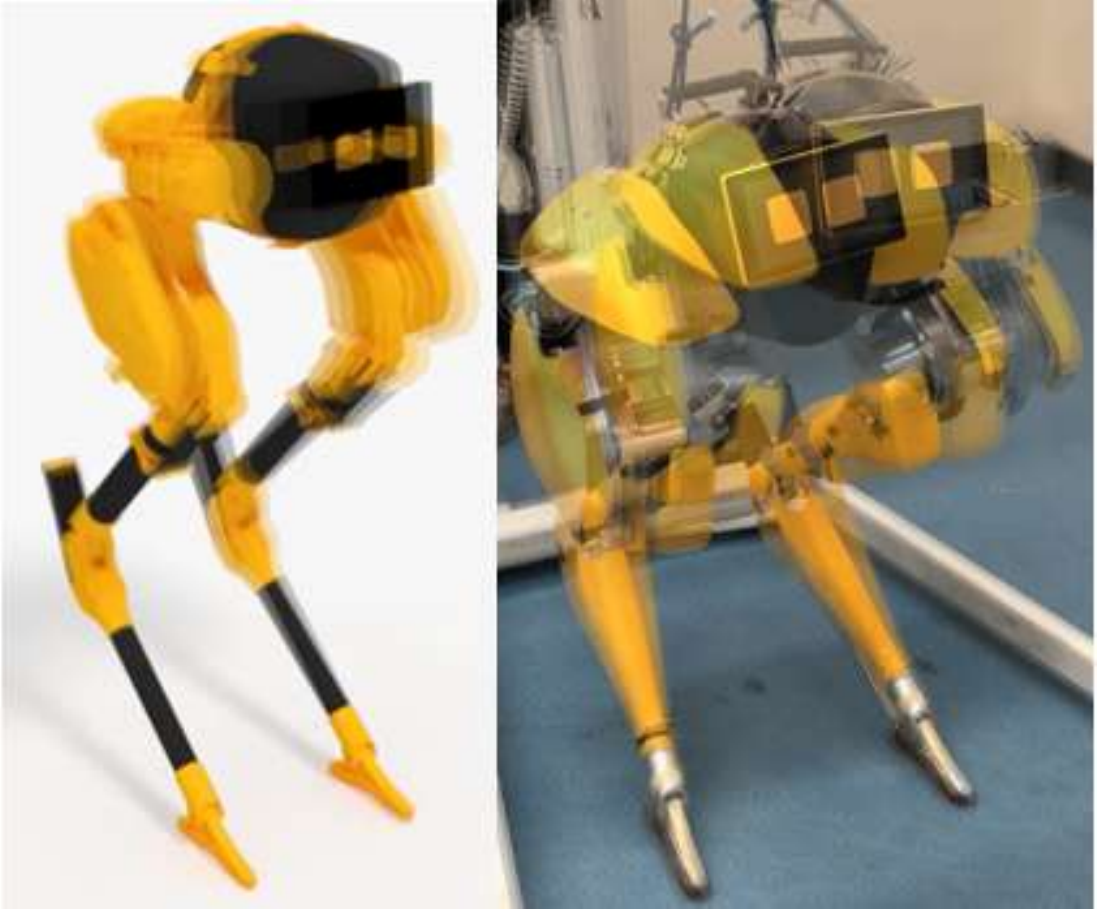}
  \caption{Tired: Shaking Head}
  \label{fig:sfig2}
\end{subfigure}\\
\hspace*{0.0em}
\begin{subfigure}{0.82\linewidth}
  \includegraphics[width=1.0\linewidth,height=0.5\linewidth]{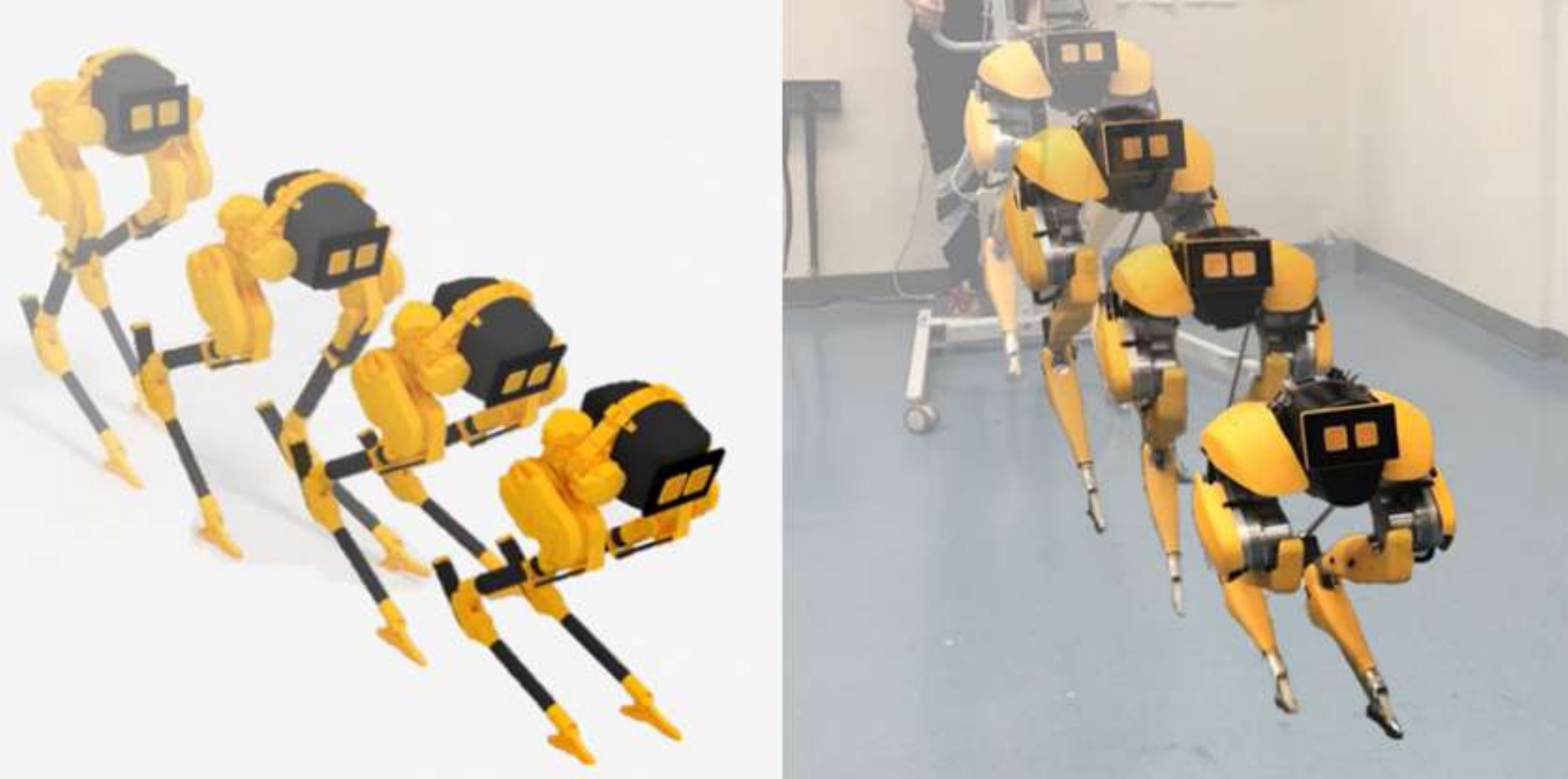}
  \caption{Curiosity: Chasing a Laser}
  \label{fig:sfig3}
\end{subfigure}

\caption{Cassie's emotive motions: the robot's emotive behaviors are animated~(left) and our proposed framework enables the dynamic robot Cassie to perform the designed motions in real time (right). Experimental video is at \url{https://youtu.be/zjfzfgb3oK8}.}
\label{fig:anim-cassie}
\end{figure}

In this paper we seek to ascertain the feasibility of utilizing a dynamic bipedal robot, Cassie Cal, to perform natural emphatic motions which ascribe a personality to a robot and enable transcending from being a ``robot'', as shown in Fig.~\ref{fig:anim-cassie}.  Cassie is a bipedal walking robot that is about one-meter high while standing and has 20 Degrees-of-Freedom~(DoF) among which are 5 actuated motors and 2 passive joints connected by leaf springs on each leg.

To address the needs for creating dynamic relatable robotic characters, three sub-problems needs to be solved:
\begin{itemize}[leftmargin=1.2cm, font=\bfseries]
    \small{
    \item[Prob.~1]Provide an animation tool to design behaviors with emotional attributes for robots. Typically, such motions are kinematic and may not be feasible for dynamic robots.
    \item[Prob.~2]Ensure the generated motions are dynamically-feasible for the robot while staying as close as possible to the original animated motions. This could be an optimization problem. 
    \item[Prob.~3]Develop a real-time controller to enable the dynamic robot to perform the optimized motions while balancing.  
    }
\end{itemize}
Effective solutions to the above-mentioned problems provide a tool to bring emotive motions to dynamic robots in real life. 

\subsection{Related Work}
For Prob.~1, algorithmic robot motion planning~\cite{latombe2012robot} is designed to execute certain tasks with given objective functions, thus resulting in inanimate motions.  To generate emotive motion for robots, there are very few solutions.  \textit{Animation creation}, as is commonly used in the entertainment industry, is one feasible approach. iCAT~\cite{IROS2004-iCAT}, followed by andyRobot~\cite{AndyRobotWeb}, Cozmo~\cite{AnkiWeb}, and Disney's Vibration-Minized Animated Character~\cite{hoshyari2019vibration}, developed animation engines to enable the robots to combine multiple interactive behaviors with believable robot animations.  Animation is a powerful tool for robotic behavior design, as animators can draw and modify robot motions interactively with visual effects~\cite{IROS2006-iCAT,takayama2011expressing}. 
However, almost all previous efforts to bridge animation and robotics has been on statically stable robot platforms with little attention focused on the dynamic properties of legged robots.  One approach widely used in humanoid-robot fields~\cite{suleiman2008human,koenemann2014real,ishiguro2017bipedal}, is imitating human motion. This method provides a tool to generate robot motion by imitating human motions, thereby addressing Prob.~1. Prob.~2 is solved by constructing a Zero-Moment-Point~(ZMP)-based optimization~\cite{vukobratovic2004zero} for the humanoid robots.  However, such a method is limited since it requires the imitated motions to be slow~\cite{koenemann2014real}, is limited to imitation and has no further creation, and is only applied on legged robots which are utilizing ZMP to stabilize.  Cassie, with line feet, cannot use such a method~\cite{vukobratovic2004zero}. 

To address Prob.~3, feedback control approaches on Cassie are introduced.  Cassie is a bipedal robot manufactured by Agility Robotics. A framework for nonlinear optimization for Cassie based on hybrid zero dynamics (HZD)~\cite{westervelt2018feedback} is constructed and solved in~\cite{hereid2018rapid}. Standing and walking controllers for Cassie using a full-model gait library are developed in~\cite{gong2019feedback}. A reduced-model-based walking controllers are designed in~\cite{xiong2018coupling,apgar2018fast}. 
However, all of the previous controllers for Cassie are unable to change its height above the ground while walking, and more importantly, there has been no effort to design emotive behavior design for Cassie.  

\subsection{Contributions}
\begin{figure}[!]
    \centering
    \includegraphics[width=0.9\linewidth]{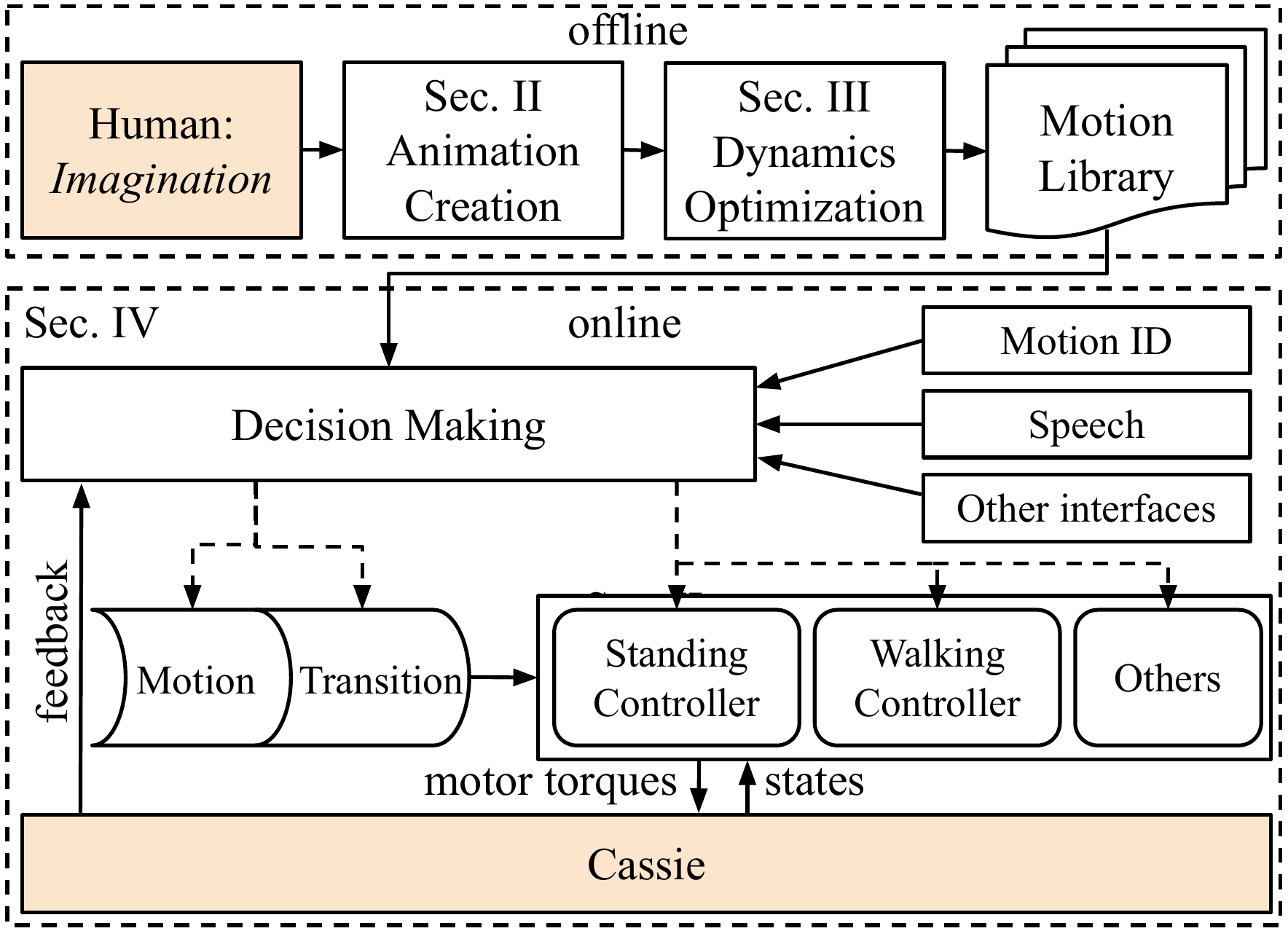}
    \caption{Framework for the dynamic relatable robotic character Cassie to behave animatedly. This pipeline enables humans to create virtual character behaviors from their imagination. After dynamics optimization, the animated motions are stored in a motion library and can later be performed by Cassie in real life.}
    \label{fig:framework}
\end{figure}

This paper makes the following contributions: 
\begin{itemize}
    \item A dynamic relatable robotic character using an animatronic bipedal robot Cassie is firstly introduced.
    \item A virtual character which agrees with holonomic constraints for Cassie is created in Blender, a 3D creation suite.   
    \item An original nonlinear constrained optimization schematic is realized to translate the animated kinematic motions to be dynamically feasible for legged robots while staying close to the animated motion.
    \item A versatile walking controller which is able to change walking height, besides the features developed in~\cite{gong2019feedback}, is developed to enable Cassie to perform agile and complex designed motions. 
    \item An end-to-end framework, as shown in Fig.~\ref{fig:framework}, is presented to enable animatronic dynamic legged robots to have personalities and emotive expressions. Experimental validation on Cassie is presented.
\end{itemize}

\subsection{Paper Structure}
The framework developed in this paper is illustrated in Fig.~\ref{fig:framework}. We first introduce animation techniques to design the behaviors for the Cassie virtual character to create expressive and emotive robot motions. This solves Prob. 1 and is presented in Sec.~\ref{sec:animation_creation}.  However, the resulting motions may not be dynamically feasible for the robot. Therefore, Sec.~\ref{sec:optimization} addresses Prob. 2 to translate the animation trajectories from the virtual character to the physical robot with dynamics constraints. Later, a motion library containing multiple story arcs is stored.  To address Prob. 3, a controller and an automaton for dynamic walking robot characters are later developed in Sec.~\ref{sec:automaton}, to enable Cassie to perform agile motions from the generated library emotively. Experiments are conducted in Sec.~\ref{sec:experiments} to validate proposed methodology and conclusions and future work is discussed in Sec.~\ref{sec:conclusion}.  

\section{Animation Creation}\label{sec:animation_creation}


Having provided an introduction and overview on our work, we now present details on how to create a Cassie virtual character and design emotive behaviors in animations for the virtual character.

\subsection{Physical Robot to Virtual Character}
\begin{figure}[!]
\centering
\begin{subfigure}{.49\linewidth}
  \centering
  \includegraphics[height=.95\linewidth]{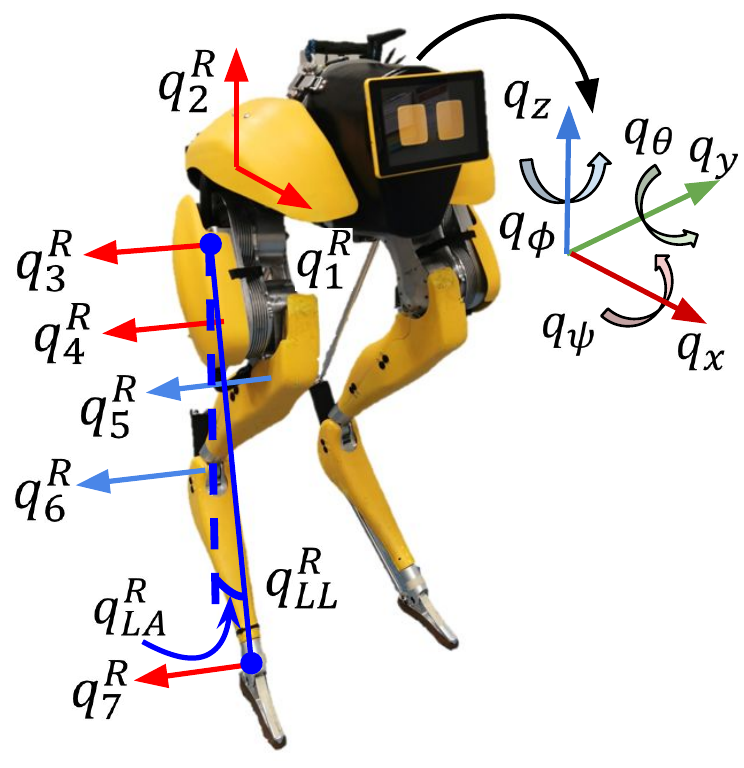}
  \caption{Cassie}
  \label{subfig:cassie_model}
\end{subfigure}\hspace*{-2em}
\begin{subfigure}{.49\linewidth}
  \centering
  \includegraphics[height=.95\linewidth]{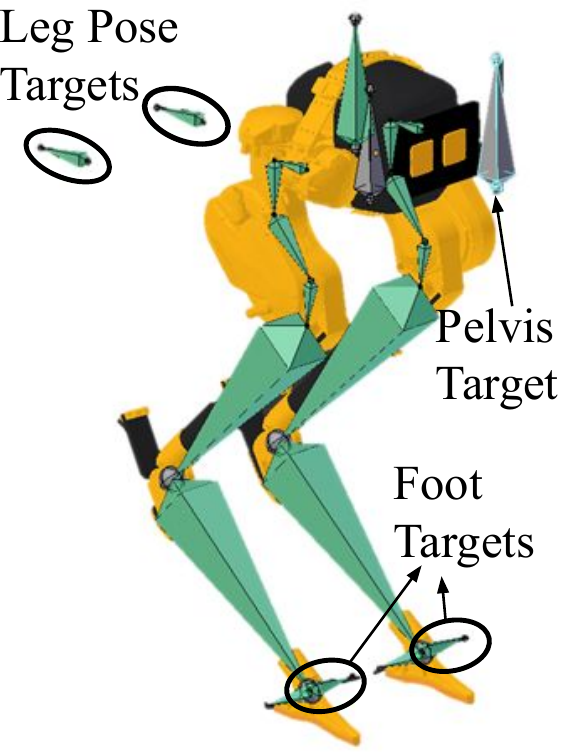}
  \caption{Cassie Virtual Character}
  \label{subfig:cassie_anim_model}
\end{subfigure}

\caption{(a) Cassie with eye-tablet in real world with the generalized floating-based coordinates and the definition of robot virtual leg length $q^{L/R}_{LL}$ and the leg angle $q^{L/R}_{LA}$. (b) Modelling and rigging of Cassie virtual character in Blender.}
\label{fig:model}
\end{figure}

Animation creation bridges animators who may have little robotics background with the robots.  Creating a virtual character for Cassie is the first step to utilize these animation techniques. In this paper, we are using Blender\textsuperscript{\textregistered} which is an open-source 3D creation suite and provides tools for modelling, rigging, animation, rendering, etc.  

In real life, Cassie, as shown in Fig.~\ref{subfig:cassie_model}, has 10 actuated rotational joints $q_{1,2,3,4,7}^{L/R}$ and 4 passive joints $q_{5,6}^{L/R}$ on its Left/Right legs. Counting the 3 transition DoF~(sagittal, lateral, horizontal) $q_{x,y,z}$ and 3 rotation DoF~(roll, pitch, yaw) $q_{\psi,\theta,\phi}$ on its pelvis, it has 20 DoF in total. To be concise, we denote the generalized coordinates of Cassie as:

\begin{align*}
\mathbf{q} = [\mathbf{q}^p,\mathbf{q}^m, q_{5}^L,q_{6}^L,q_{5}^R, q_{6}^R]^T
\end{align*}
where
\begin{align*}
\mathbf{q}^p &= [q_{x}, q_{y}, q_{z}, q_{\psi}, q_{\theta}, q_{\phi}]^T,\\
\mathbf{q}^m &= [q_{1}^L, q_{2}^L, q_{3}^L, q_{4}^L, q_{7}^L, q_{1}^R, q_{2}^R, q_{3}^R, q_{4}^R, q_{7}^R]^T,
\end{align*}
and $\mathbf{q}^p$, $\mathbf{q}^m$ denote pelvis DoFs and actuated motors positions, respectively. 

We also denote $\mathbf{q}_i = [q_{i}^L, q_{i}^R]^T$ as the $i$-th joints on both legs, and $\mathbf{q}_{LL}=[q^L_{LL}(q^L_3, q^L_4), q^R_{LL}(q^R_3, q^R_4)]^T$ and $\mathbf{q}_{LA}=[q^L_{LA}(q^L_3, q^L_4), q^R_{LA}(q^R_3, q^R_4)]^T$ as virtual leg length and leg pitch, respectively. Moreover, we define $\mathbf{x}^{L/R}_{foot}(\mathbf{q})=[x^{L/R}_{foot}, y^{L/R}_{foot}, z^{L/R}_{foot}]^T$, the Cartesian positions of robot's end-effectors which are the feet on the Left/Right legs. We denote $\mathbf{x}_{CoM}(\mathbf{q})=[x_{CoM}, y_{CoM}, z_{CoM}]^T$ as Cassie's Center-of-Mass~(CoM) Cartesian position. These terms can be determined by forward kinematics.

Based on this, we created a rigged character model of Cassie in the virtual world, as illustrated in Fig.~\ref{subfig:cassie_anim_model}. Rigging involves taking the geometric information from each link on Cassie and mapping the rigid body as a ``bone''~(green link in Fig.~\ref{subfig:cassie_anim_model}) and kinematic connection as ``joint'' which connects two adjacent bones.  While animators can design a motion by specifying sequences of joint positions, it is more efficient and natural for them to specify the poses of end effectors (either the feet on the ground or the pelvis). Towards this, we added target bones such as a \textit{Pelvis Target (PT)} which determines the pelvis pose, \textit{Leg Pose Targets~(LPTs)} to change leg's orientation, and \textit{Foot Targets (FTs)} to position two feet, as presented in Fig.~\ref{subfig:cassie_anim_model}.  

\subsection{Inverse Kinematics Constraints}\label{subsec:ik}
Once animators have defined the pose of the target bones, like placed Foot Targets \textit{(FTs)} on the ground and tilted Pelvis Target \textit{(PT)} to make Cassie look in some direction, the rest of joint positions can be determined by solving an inverse kinematics~(IK) problem. However, as mentioned, two joints corresponding to the joints from knee to shin $\mathbf{q}_{5}$ and the ankle joint $\mathbf{q}_{6}$ are passive and are connected through stiff leaf springs. When the springs are uncompressed, there exists additional constraints related to the knee and ankle joints, specifically, $\mathbf{q}_5\equiv0$, $\mathbf{q}_6 + \mathbf{q}_4 - 13^{\circ} = 0$~\cite{gong2019feedback}. 

Therefore, the inverse kinematics problem can be formulated as:
\begin{align*}
    \min_{\mathbf{q}} \quad &\|\mathbf{x}^{L/R}_{foot}(\mathbf{q}) - \mathbf{x}(FTs)\|_2^2 \tag{1}\label{eq:ik}\\
    \textrm{s.t.} \quad & \mathbf{q}_{min} \leq \mathbf{q} \leq \mathbf{q}_{max}, \\
             & \mathbf{q}^p - p(PT) = 0, \\
             & \mathbf{q}_2-f(LPTs) = 0, \\
             & \mathbf{q}_5 = 0, \\
             & \mathbf{q}_6 + \mathbf{q}_4 - 13^{\circ} = 0, \\ 
\end{align*}
where the objective function is the $L_2$ distance between the Cartesian positions of the robot's feet~$\mathbf{x}^{L/R}_{foot}$ and Foot Targets~$\mathbf{x}(FTs)$. Pelvis pose $\mathbf{q}^p$ is constrained to follow the pose of Pelvis Target ${p(PT)}$. $f(LPTs)$ captures the relative rotation between the Leg Pose Target (\textit{LPT}) to its corresponding leg orientation in that leg's horizontal planes. The constraint of $\mathbf{q}_2-f(LPTs) = 0$ enforces each leg to always point towards its \textit{LPT}.

As a result, joint trajectories which satisfy Cassie's kinematic constraints are obtained by solving optimization problem \eqref{eq:ik} during the animation creation. We develop a custom solver as a Python plugin for Blender as its IK solver does not support internal joint constraints. 
 
\subsection{Creating Animation}

\begin{figure}[t]
\centering
\begin{subfigure}{.8\linewidth}
  \centering
  \includegraphics[width=1\linewidth]{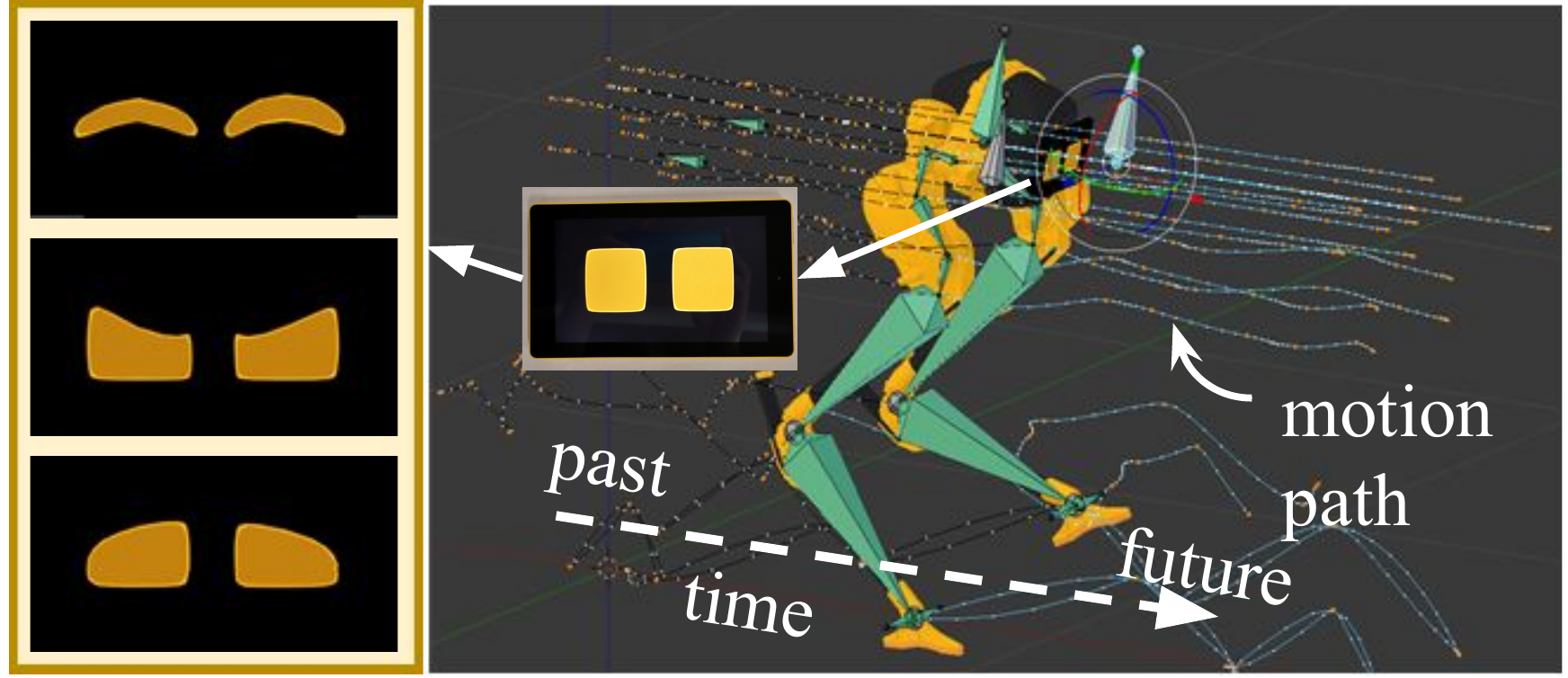}
  \caption{Eyes Expressions and Animation Sketching}
  \label{subfig:walking_animation}
\end{subfigure}
\begin{subfigure}{.8\linewidth}
  \centering
  \includegraphics[width=1\linewidth]{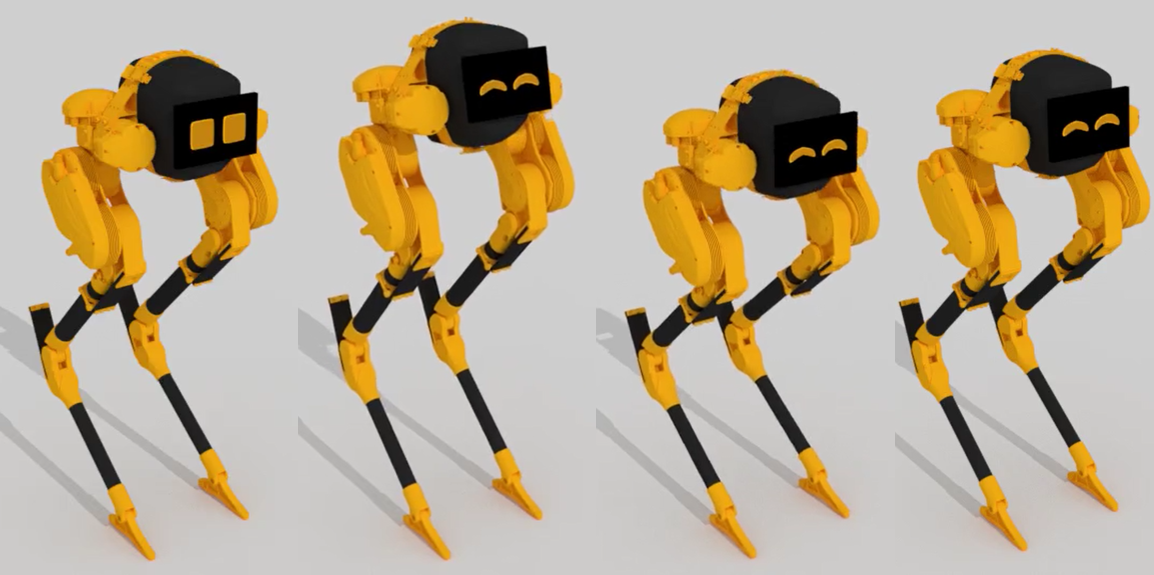}
  \caption{Rendered Happy Motion}
  \label{subfig:happy_animation}
\end{subfigure}

\caption{Animation of the Cassie virtual character. (a) Walking animation where the  motion path presents the trajectories of each joint of the character along time. On the left hand side shows examples  of  visual  eye  expressions (from top to bottom: happy, angry, sad). (b) The designed motion, such as happy motion, can be later rendered and produce an animation.}
\label{fig:animation}
\end{figure}

To provide visual cues of emotion for the character, we also added eyes for the virtual character as illustrated in Fig.~\ref{subfig:walking_animation}. The eye motion can be designed jointly with the body motion. 
In real life, we added a tablet to display visual eye expressions as shown in Fig.~\ref{subfig:cassie_model}. 
Once we have a satisfying virtual character, animators could be given free rein to design emotive and expressive behaviors for the character. Animation technique allows the animator to focus on the creation of motion through creation of \textit{keyframes} with the poses in the frames between two adjacent keyframes interpolated automatically. The animator is free to choose different methods of interpolation, such as B\'{e}zier or linear, provided by Blender. A typical walking animated motion is shown in Fig.~\ref{subfig:walking_animation}. Please note that the keyframe is more heavily weighted than the interpolated frame in terms of motion design in the animation field. This property will be used in the following optimization in Sec.~\ref{sec:optimization}.  

Once an animated motion is completed, the animator is able to utilize the render engine in Blender to create a professional animated movie, like Fig.~\ref{subfig:happy_animation}. Moreover, we provide a tool for animators to specify the mode, \textit{i.e.} standing or walking, of the designed motion, and to extract and store all joint trajectories in the animation they created. We denote the animation as $\mathcal{Q}_{ani}$ 
with sequence of frames $\{\mathbf{q}_{ani}\}$.


\section{Animation-to-Dynamics Optimization} \label{sec:optimization}
Having created a pipeline to create animation behaviors for Cassie, we will next see how to transfer these to hardware. While we included an IK optimization in Sec.~\ref{subsec:ik} to ensure the animated motion is kinematically feasible for Cassie, however, the created motion may still may violate certain laws of physics. 
Therefore, the objective of this section is to optimize for a trajectory that is dynamically-feasible, enforced by various constraints, while also being close to the animation. 

We propose to solve this by formulating a direct-collocation based nonlinear optimization. However, the character's motion could either be in double-support standing or could be walking, wherein the dynamics constraints are different. Moreover, the input animated motion can be arbitrarily long, \textit{i.e.}, could even be tens of seconds (resulting in thousands of collocation nodes), and direct-collocation at that large time spans is computationally hard to optimize. Therefore, we address this by firstly separating the optimization problem into standing and walking scenarios where distinct optimization solutions are discussed, and secondly by splitting the optimization into shorter time spans with continuity constants between each.

\subsection{Optimization for Standing}

\begin{figure}[t]
    \centering
    \includegraphics[width=0.9\linewidth]{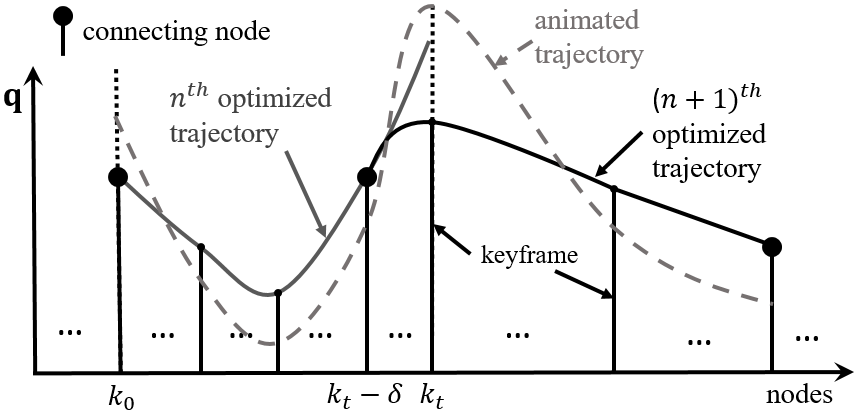}
    \caption{Schematic for animation-to-dynamics optimization in the standing scenario. During optimization, the arbitrarily-long animated trajectory is broken into sub-motions and optimization is applied to each smaller trajectory in sequence. Between each sub-motion, such as the one from $k_0$-th node to $(k_t-\delta)$-th node, smoothness constraints are applied on the connecting node.}
    \label{fig:standing_optimization}
\end{figure}

The input animated motion $\mathcal{Q}_{ani}$ is a sequence of nodes $\{\mathbf{q}_{ani}\}$ wherein each node is the full state of the animated Cassie character with a timestamp. As the input animation has arbitrarily long frames, 
the sequence $\mathcal{Q}_{ani}$ elapsed in seconds could contain thousands of nodes. In order to make the optimization problem solvable, we break the optimization into multiple smaller ones between each sub-motion while maintaining the smoothness of the entire motion, as illustrated in Fig.~\ref{fig:standing_optimization} and explained in Algorithm~\ref{algorithm:optimization_standing}. 

\begin{algorithm}[t]
\caption{Optimization for Standing Motion}
\begin{algorithmic}[1]
\label{algorithm:optimization_standing}
\begin{small}
 \renewcommand{\algorithmicrequire}{\textbf{Input:}}
 \renewcommand{\algorithmicensure}{\textbf{Output:}}
 \REQUIRE Animated Motion $\mathcal{Q}_{ani} = \{\mathbf{q}_{ani}(k)\}$, $k=1,2,\dots,N$
 \ENSURE  Optimized Motion $\mathcal{Q}_{opt} = \{\mathbf{q}_{opt}(k)\}$, $k=1,2,\dots,N$
 \\ \textit{Initialization}: 
 $(k_{0},k_{t})\leftarrow (1,1+n)$, $\mathcal{Q}_{opt}=\varnothing$

    \WHILE{$k_{t} \leq N$} 
        \STATE 
        $\{\mathbf{q}_{opt}\}\leftarrow$ solution of: \\
        $\min_{\{\mathbf{q}\}}\sum_{i=k_{0}}^{k_{t}} ||\mathbf{u}(i)||_2^2+ c||\mathbf{q}_{ani}(i)-\mathbf{q}(i)||_2^2$ \hfill(2)\label{eq:standing_opt}\\

        $\quad \textrm{s.t.} 
        \quad t_{ani} \leq t_{opt} \leq 4 t_{ani}$ \hfill(2.1)\\
        $\qquad\quad \mathbf{q}_{min}\leq \mathbf{q}\leq \mathbf{q}_{max}$ \hfill(2.2)\\
        $\qquad\quad \mathbf{u}_{min} \leq \mathbf{u} \leq \mathbf{u}_{max}$ \hfill(2.3) \\ 
        $\qquad\quad D(\mathbf{q})\ddot{\mathbf{q}} + C(\mathbf{q},\dot{\mathbf{q}})\dot{\mathbf{q}} + G(\mathbf{q}) = B\mathbf{u} + \lambda J^T(\mathbf{q})$\hfill(2.4)\\
        $\qquad\quad J(\mathbf{q})\ddot{\mathbf{q}} + \dot{J}(\mathbf{q},\dot{\mathbf{q}})\dot{\mathbf{q}} = 0$ \hfill(2.5)\\
        $\qquad\quad \mathbf{x}_{foot}^R(\mathbf{q}) = [0,0,0]^T$ \hfill(2.6)\\ 
        $\qquad\quad [-10,-10,0]^T \leq \mathbf{x}_{foot}^L(\mathbf{q}) \leq [10,10,0]^T$ \hfill(2.7)\\
        $\qquad\quad F_{c}(\mathbf{q},\dot{\mathbf{q}},\ddot{\mathbf{q}})$ stays in the friction cone~$(\mu=0.6)$ \hfill(2.8)\\
        $\qquad\quad \mathbf{x}_{CoM}(\mathbf{q})$ lies in the support region  \hfill(2.9) \\
        $\qquad\quad ||\mathbf{q}^p_{ani}(k_0) - \mathbf{q}^p(k_{0})||_2^2 = 0$ \hfill(2.10)\\
        $\qquad\quad ||\dot{\mathbf{q}}^p_{ani}(k_0) - \dot{\mathbf{q}}^p(k_{0})||_2^2 = 0$ \hfill(2.11)\\
        $\qquad\quad ||\dot{q}(k_{t})_{x,y,z}||_2 = 0$ if $k_{t} = N$ \hfill(2.12)\\
        
        \STATE        
        $(k_{0},k_{t})\leftarrow (k_{t}-\delta, k_{t}-\delta+n)$
        \STATE 
        $\mathcal{Q}_{opt} = \mathcal{Q}_{opt} \cup \{\mathbf{q}_{opt}\}$
    \ENDWHILE

 \RETURN $\mathcal{Q}_{opt}$ 
\end{small}
\end{algorithmic}
\end{algorithm}

In Algorithm~\ref{algorithm:optimization_standing}, we cut the long sequence of nodes whose horizon is $N$ into sub-motions which have $n$ nodes for each. As the keyframe is more important than the interpolated frame in the animation, we want the start and end nodes to be keyframes, and to have similar numbers of keyframes among sub-motions. Therefore, the value of $n$ is different in each sub-motion. On each sub-motion, the optimization problem formulated as~\eqref{eq:standing_opt} in Algorithm~\ref{algorithm:optimization_standing} is solved. The objective function~\eqref{eq:standing_opt} is set to minimize the sum of energy consumption and the $L_2$ distance between the optimized poses $\mathbf{q}$ to the animated ones $\mathbf{q}_{ani}$ in the sub-horizon. The parameter $c$ has three major roles: to weigh between the similarity to the animated motion and total energy consumption, to emphasize the poses of the animation keyframe in the optimized motion, and to push the end of the sub-motion to get as close as possible to the animated one. Therefore, $c$ is set to $1500$, $10^4$, $10^5$ at normal frames, keyframes, and the ending frame, respectively.  

Besides putting boundaries on the elapsed time~(2.1), states~(2.2), and inputs~(2.3), (2.4) enforces Cassie's dynamics while (2.5-2.7) formulate the kinematics constraints and place the left and right foot Cartesian positions $\mathbf{x}^{L/R}_{foot}$ on the ground. Moreover, (2.8) prevents the feet slipping on the ground whose static friction ratio $\mu$ is 0.6, and (2.9) balances the robot by limiting the CoM position $\mathbf{x}_{CoM}$ projected on the ground lying in the support region composited by two feet $\mathbf{x}^{L/R}_{foot}$ as formulated in~\cite{hereid2018rapid}.

Furthermore, in order to connect two adjacent sub-motions, we added (2.10) to enforce the start node of current sub-motion to overlap with the end node of the last subsequent motion, while adding velocity constraints (2.11) to smooth the connection, as illustrated in Fig.~\ref{fig:standing_optimization}. Please note that we only restrict the smoothness of the pelvis motion $\mathbf{q}^p$ and let the kinematics constraints solve the rest of joints. Also, to leverage the smoothness constraints, we begin from $\delta$ number of frames ahead of the end node of the previous sub-motion when sliding through the entire motion. In practice, $\delta$ varies to ensure each sub-motion starts with a keyframe. Note that (2.12) is also added for zero-transition-velocity of the pelvis at the ending of the entire motion.

Therefore, in this way, we can obtain optimized motions with arbitrary time spans as the algorithm only focuses on a short but optimize-able long motion and the connection points in between.   

 \begin{figure*}[t]
    \centering
    \begin{subfigure}{0.45\textwidth}
        \centering
        \includegraphics[height=.4\linewidth]{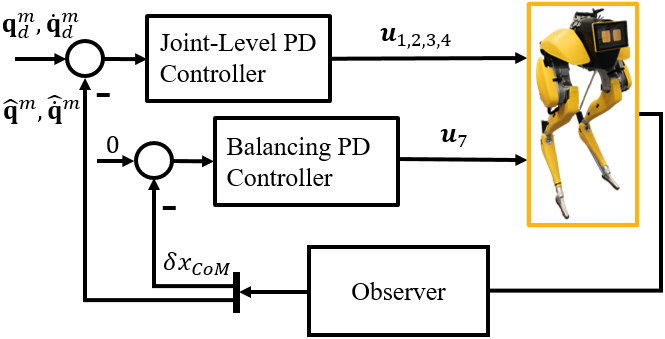}
        \caption{Standing Controller} \label{fig:standing_controller}
    \end{subfigure} \hspace*{-2em}
    \begin{subfigure}{0.45\textwidth}
        \centering
        \includegraphics[height=.4\linewidth]{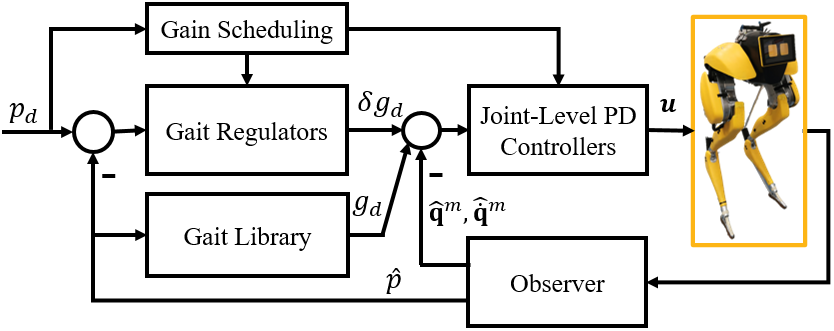}
        \caption{Walking Controller} \label{fig:walking_controller}
    \end{subfigure} 
    \caption{Control structure for animated Cassie. (a) The input to the standing controller is the joints trajectories from the motion library. (b) The walking controller is designed to track input gait parameter $p_d$ which is the sagittal and lateral velocity and desired walking height. It interpolates the gait library by observed gait parameter $\hat{p}$ to obtain a reference gait $g_d$ and regulate this by a gait regulator. The input walking height is also fed forward to a gain scheduler which changes the parameters used in the controllers.}
\end{figure*}

\subsection{Optimization for Walking} \label{subsec:walking_optimization}
For a walking robot, a \underline{gait} $g$ is defined as periodic joint position and velocity profiles for all actuated joints. For each gait, there is a \underline{gait parameter} $p$ that identifies it. It is hard to require Cassie to mimic whole body motion while maintaining stable walking gaits using a real-time controller. Therefore, we mediate the motion imitation and nonlinear optimization by limiting the imitated DoF to the gait parameter which consists of sagittal and lateral walking velocities, $v_x$ and $v_y$, and walking height $h$ (the height of pelvis above ground) from the animation, \textit{i.e.}, $p=[v_x,v_y,h]^T$. In this way, we choose to design a library of periodic walking gaits for different gait parameters, and let Cassie track the animated ones. 

The optimization is applied on a hybrid model which contains two continuous domains, \emph{right stance} and \emph{left stance}, and two transition domains called \emph{left impact} and \emph{right impact}. During each stance domain, one of the legs is pinned on the ground. Domain transitions are triggered at foot impact events causing the model to transit from \emph{right stance} to \emph{left stance} and \emph{left stance} to \emph{right stance}, respectively. Periodic walking motion can be realized by starting and ending in \emph{right stance}. 

The optimization formulation to generate walking gaits for Cassie is described in~\cite{gong2019feedback}. It includes constraints on average velocity to match the given gait parameter, virtual constraints to have hybrid zero dynamics which is impact-invariant~\cite{westervelt2018feedback}, and others. For solving for gaits with a desired walking height $h$ for Cassie, we include two more constraints:
\begin{itemize}
    \item Pelvis height: $q_z \in [h-0.035,h+0.035]$~m   
    \item Step time: $t \in [t_{min}(h),0.4]$, where $t_{min}(h)$ is linearly interpolated from 0.2s to 0.4s with respect to $h$
\end{itemize}

In this way, we generated a \underline{gait library} $\mathcal{G}$ for the following gait parameters $p$:
\begin{itemize}
    \item 11 forward/backward walking speeds $v_x$ in the range $[-1,~1]$~m/s
    \item 11 left/right side-stepping speeds $v_y$ in the range $[-0.3,~0.3]$~m/s
    \item 11 walking heights $h$ in the range $[0.65,~1]$~m per step
\end{itemize}

This gait library has $11
\!\times\!11\!\times\!11\!\!=\!\!1331$ gaits parametrized by B\'{e}zier polynomials and consists of a set $\mathcal{G} = \{g_i(t_i,\mathbf{\alpha}^j_i)\}$ where $\alpha^j_i$ represents the coefficients of the B\'{e}zier polynomial for the $j$-th motor in the $i$-th gait $g_i$ and $t_i$ encodes the period for the gait. 

\subsection{Motion Library}
Based on the animation creation process, we created multiple animations with different emotions (happy, sad, tired, etc.). For each animated motion, the above-proposed optimizations are applied to obtain dynamically feasible motion for Cassie that is close to the animated motion. Also, a tag indicating the motion's emotion, as well as the control mode, \textit{i.e.} standing or walking, is added. In this way, we store the optimized motions in a motion library from which they can be retrieved by their unique IDs. 
\section{Control and Automaton for Robot Character}\label{sec:automaton}

Once a library of dynamically feasible motions is obtained, we need a real-time controller for Cassie to replay the input motions. Moreover, switching between different controllers, \textit{e.g.} from standing to walking, or performing different motions in sequence, could destabilize the robot if the transition happens inappropriately. Therefore, a high-level automaton is needed to decide which motion to send to the controller, and when and how the transition should start and end.

\subsection{Versatile Standing and Walking Controller for Cassie} \label{subsec:controllers}
Basic standing and walking controllers for Cassie are developed in~\cite{gong2019feedback}. However, 
the walking controller is not able to reliably track the velocity commands in the lateral direction and at different walking heights. Therefore, based on the previous work, we developed a standing controller, as shown in Fig.~\ref{fig:standing_controller} which is capable of tracking all 6 DoF motions for the pelvis, as well as a controller that is able to significantly change body height while walking in 3D space, as illustrated in Fig.~\ref{fig:walking_controller}.

\subsubsection{Standing Controller} The reference $\mathbf{q}^m_{d}$ and $\mathbf{\dot{q}}^m_{d}$ is the motions of all the motors from the motion library. However, the foot joint positions do not track the animated motion but rather help in keeping the feet on the ground and in rejecting perturbation to keep the sagittal component of the Center-of-Mass $\mathbf{x}_{CoM}$ to stay in the support region, \textit{i.e.}, $\delta x_{CoM} = ||\mathbf{x}_{CoM}(1) - \frac{1}{2}(\mathbf{x}^R_{foot}(1) + \mathbf{x}^L_{foot}(1)) ||_2 \equiv 0$. 
For every actuated joint, a PD controller is applied to track the reference joint trajectories based on the feedback $\mathbf{\hat{q}}^m$, $\hat{\mathbf{\dot{\mathbf{q}}}}^m$ from the encoders.

\subsubsection{Walking Controller} A desired gait $g_d$ is obtained by linearly interpolating the gait library $\mathcal{G}$ obtained in Sec.~\ref{subsec:walking_optimization} with respect to the observed current robot's gait parameters $\hat{p} = [\hat{v}_x,\hat{v}_y,\hat{h}]^T$.
To maintain a stable walking gait and track the reference gait parameter $p_d$, a gait regulator is applied to the desired gait. 
The gait regulator has a correction $\delta g_d$ with three terms defined as:
\begin{equation}\tag{3}
    \!\delta g_d {=}\!
    \begin{bmatrix}
    \delta q^{sw}_3\\ \delta q^{sw}_{1} \\ \delta q^{st}_4 
    \end{bmatrix}\!{=} K_P(h^d) (p^d - p) + K_D(h^d) (\dot{p}_d - \hat{p}')
\end{equation}
where $\delta q^{sw}_{LA}$, $\delta q^{sw}_{1}$ are corrections for the leg pitch and abduction joint of the \emph{swing} leg, and $\delta q^{st}_{LL}$ represents the correction to the \emph{stance} leg length. Also, $p^d$ and $\hat{p}$ are the desired and current gait parameters, while $\hat{p}'$ is the numerical differentiation of $\hat{p}$. 
Also, we map the walking height $h$ to the stance leg length $q^{st}_{LL}$ with a constant offset. 

In this way, the commanded gait becomes $g_d + \delta g_d$ which is the sum of the reference gait $g_d$ and regulating term $\delta g_d$, with proper dimension. 
Later, joint-level PD controllers applies input torque commands to each motor, based on the measured joint states $\mathbf{\hat{q}}^m$, $\hat{\mathbf{\dot{\mathbf{q}}}}^m$. 

This cascade walking controller has a gain scheduling module wherein the PD gains are scheduled based on the reference walking height.
This is achieved by linearly interpolating in a look-up parameter table with respect to desired walking height $h_d$. 

\subsection{Automaton for Robot Character}

\begin{figure}[!]
\centering
\begin{subfigure}{.8\linewidth}
  \centering
  \includegraphics[width=.7\linewidth]{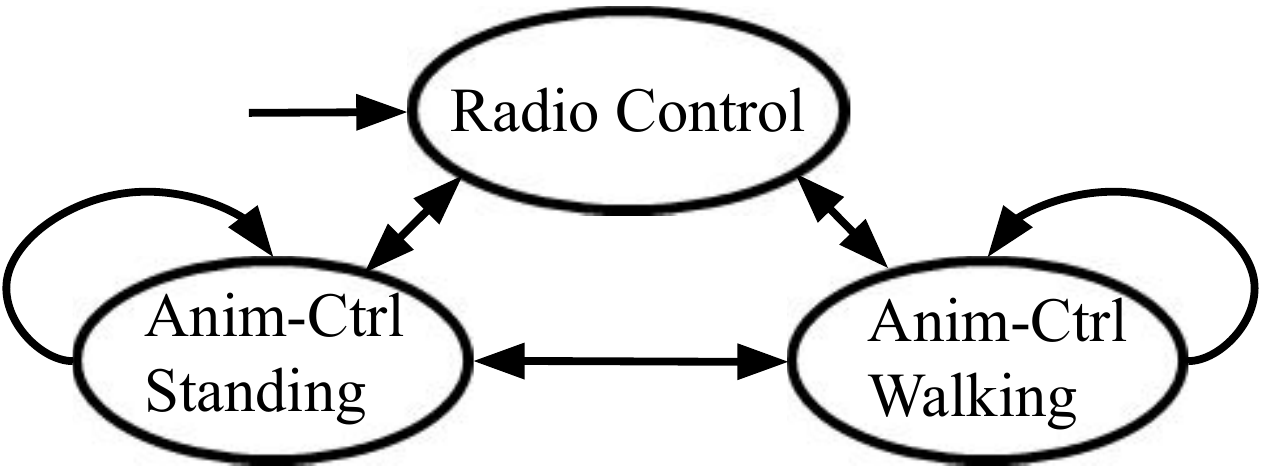}
  \caption{State Machine}
  \label{subfig:state_machine}
\end{subfigure} \\ 
\begin{subfigure}{0.95\linewidth}
  \includegraphics[width=.95\linewidth]{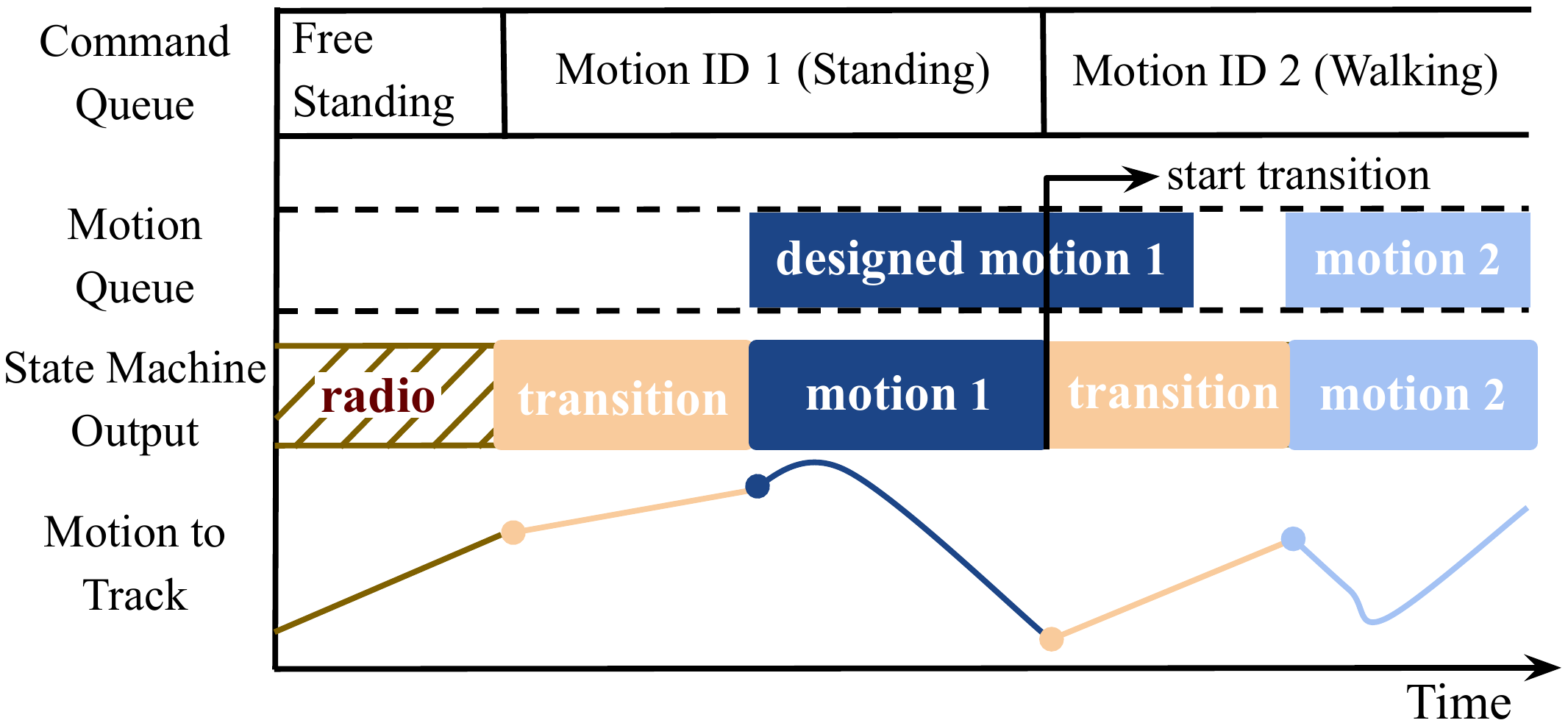}
  \caption{Online Motion Blending Schematic}
  \label{subfig:automaton_result}
\end{subfigure}

\caption{Automaton for robot character. (a) A state machine is running in real time to concatenate and smooth the reference motions. (b) The output from the state machine is a blended motion with motions from motion library or radio commands and transitions in between. Such motion is sent to controller to track.}
\label{fig:automaton}
\end{figure}

When we splice multiple motions together, there is no guarantee that the connecting point is continuous when we switch from one motion to the next. Such discontinuity may cause the robot to fall.
Also, to ensure safety, radio commands are needed for human operators to take over the control of the robot. Switching between different animated motions, as well as changing between animation-controlled mode and manual radio-controlled mode, may drive the robot to fall and thus such discontinuity should be smoothed in real time. To address this problem, we propose an online automaton, as shown in Fig.~\ref{subfig:state_machine}, that is capable of concatenating two motions and producing a smooth transition accounting for robot's current state. 

Note that the radio can command Cassie to stand or walk, and each motion comes with a control mode. Therefore, the transition types between adjacent states are within the Cartesian product of the modes of \{\textit{Standing}, \textit{Walking}\} and \{\textit{Animation-Controlled}, \textit{Radio-Controlled}\}. The four elemental types of transitions for the animatronic walking automaton and the transitions can be defined as: 
\begin{itemize}
    \item \textit{Stand to Walk}: Change the standing height to the desired walking height, then lean Cassie's pelvis to left and lift its right leg to start walking. 
    \item \textit{Walk to Stand}: Decelerate robot's velocity to zero and switch to standing mode when a step is completed. 
    \item \textit{Radio from/to Animation}: Transit from current states to the desired states which are the first frame in animation or commands from the radio. The state trajectories are obtained by linear interpolation. 
    \item \textit{Animation to Animation}: This transition happens when the motion ID is changed. States are commanded to travel by linearly interpolating current state and the first frame of the next motion. 
\end{itemize}
As a result, the transitions in the state machine shown in Fig.~\ref{subfig:state_machine} can be determined by combining the elemental transitions pairwise.

As suggested in Fig.~\ref{fig:framework} and Fig.~\ref{subfig:automaton_result}, the real-time automaton for the animatronic walking robot works as follows: 1) a motion ID, determined by a queue of motion IDs, or by requests from the interacting human, is sent to the state machine in Fig.~\ref{subfig:state_machine}; 2) the state machine decides whether it needs a transition, what kind of transition should take place, and when should the designed motion start, based on the real-time feedback from the robot, radio commands, and input motion ID; and 3) the motion to track is then composed by the transitions and interpolating the designed motion and is sent to the real-time controller for Cassie to track.  

In this way, an end-to-end framework for the animatronic walking robot Cassie is built up and the requirements for Cassie's dynamically feasible expressive motions are met. 

\section{Results and Experiments} \label{sec:experiments}

\subsection{Optimization Results} 

\begin{figure}[t]
\centering
\begin{subfigure}{.475\linewidth}
  \includegraphics[width=\linewidth]{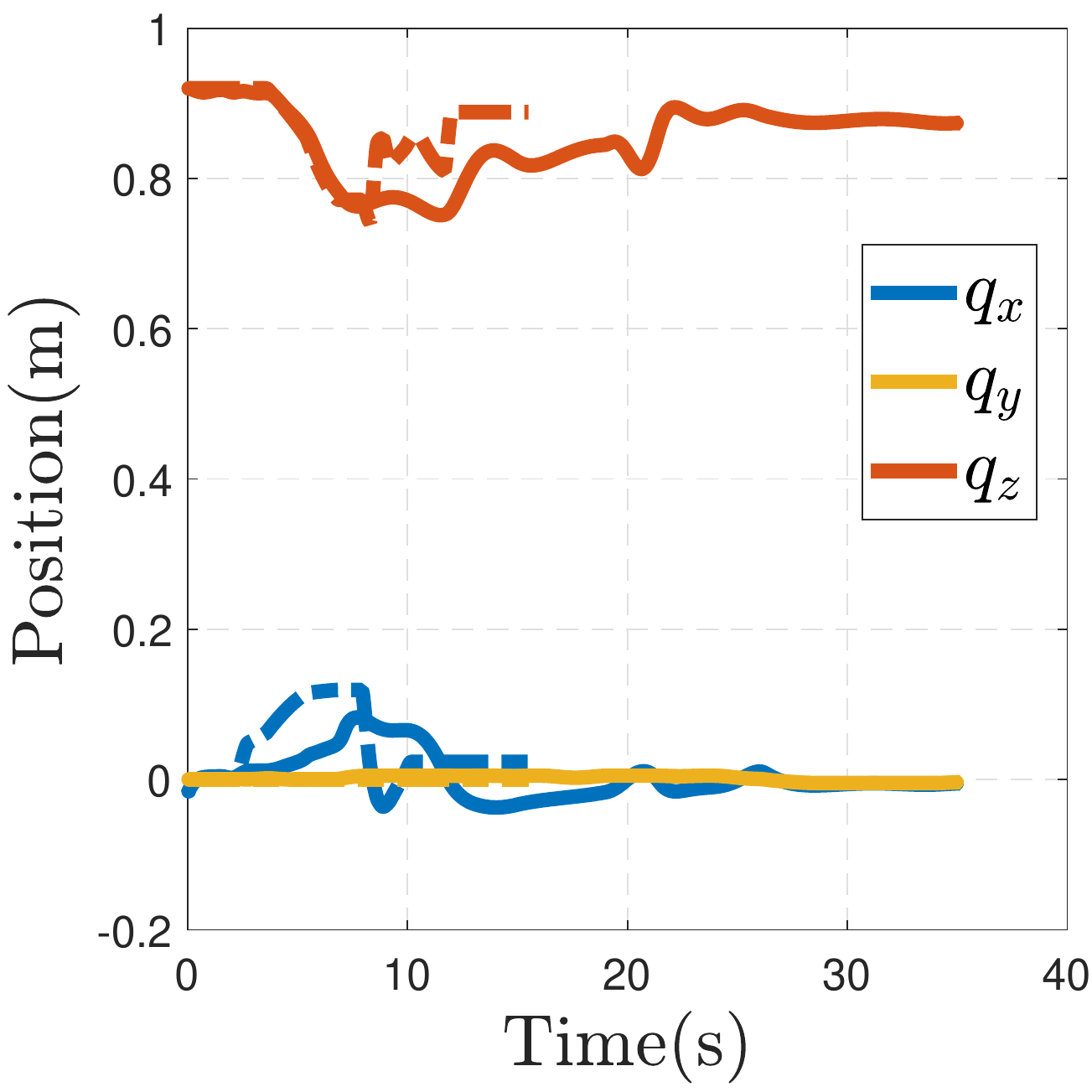}
  \caption{}
  \label{subfig:stand_opt_loc}
\end{subfigure}
\begin{subfigure}{.475\linewidth}
  \includegraphics[width=\linewidth]{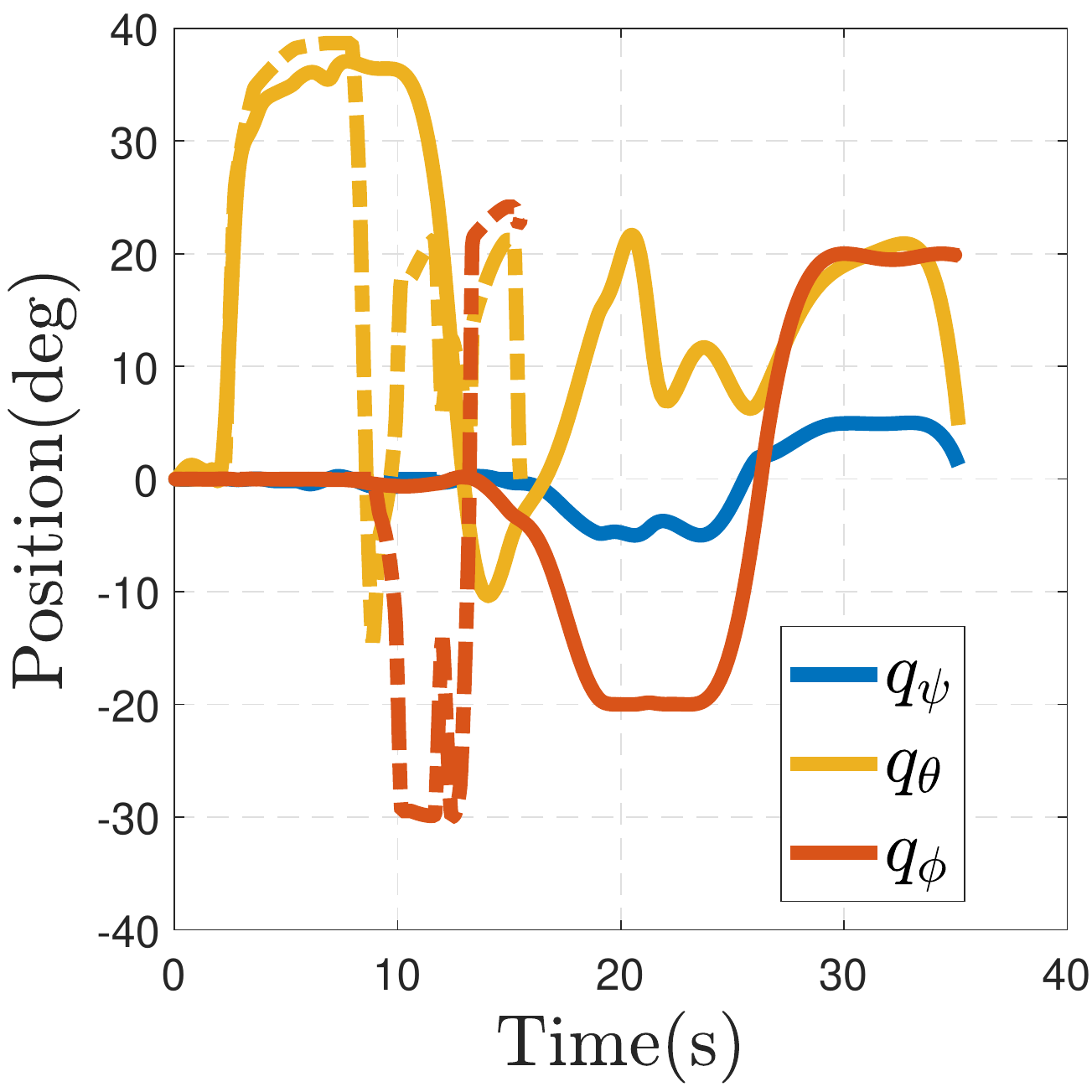}
  \caption{}
  \label{subfig:stand_opt_rot}
\end{subfigure}\\
\begin{subfigure}{.4\linewidth}
  \includegraphics[width=\linewidth]{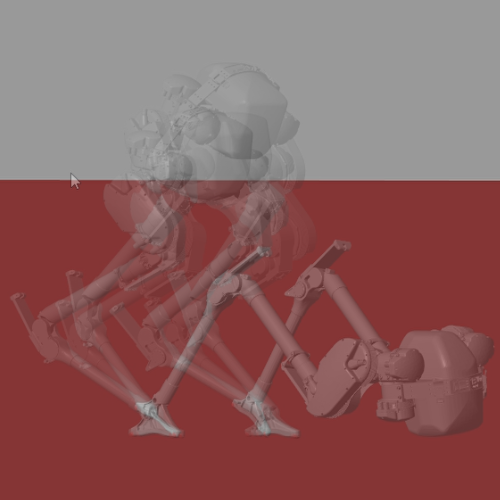}
  \caption{}
  \label{subfig:ani_stand_motion}
\end{subfigure}\hspace*{1.5em}
\begin{subfigure}{.4\linewidth}
  \includegraphics[width=\linewidth]{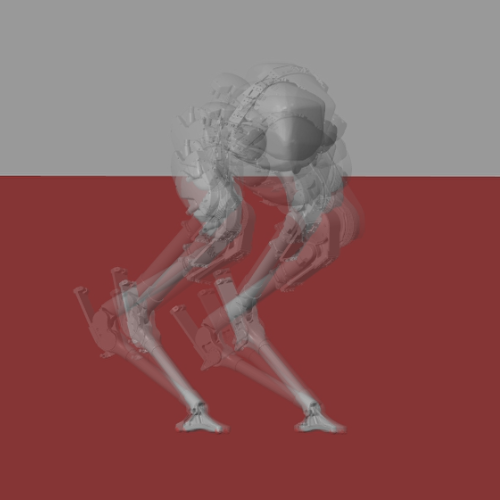}
  \caption{}
  \label{subfig:opt_stand_motion}
\end{subfigure}

\caption{Results of the optimization for standing. (a)(b)  Comparisons of resulted pelvis transitional and rotational position $\mathbf{q}^p$ profiles from animation and optimization. The motion from animation is drew in dash line while its optimized motion is presented in solid line. (c) Dynamic simulation of Cassie with the animated motion results in a fall. (d) No failure happens after the animated motion is optimized using the same controller and dynamic simulator.}
\label{fig:opt_results}
\end{figure}

We formulated the optimization problems in Sec.~\ref{sec:optimization} by using the open-sourced optimization toolkit CFROST~\cite{hereid2018rapid} and solved by IPOPT~\cite{biegler2009large}. The proposed standing optimization scheme shows the capacity to obtain dynamically feasible motion while minimizing the deviation from the animated trajectories for animations with any duration. A 5-second standing motion can be solved within an hour offline, using the full-order model of the robot. As a test case we compare the animated motion and optimized motion on the same standing controller introduced in Sec.~\ref{subsec:controllers} in a high-fidelity physical simulator of Cassie built in Simulink. The result is demonstrated in Fig.~\ref{fig:opt_results}. It shows that the animated motion, shown as a dash line in Fig.~\ref{subfig:stand_opt_loc},\ref{subfig:stand_opt_rot} causes Cassie to fall as illustrated in Fig.~\ref{subfig:ani_stand_motion}. This is because the animated motion changes so fast, \textit{i.e.} pelvis yaw $q_{\phi}$ changes over 50 degrees within 1 second at 13s, which is beyond what the robot can achieve. In contrast, in Fig.~\ref{subfig:opt_stand_motion}, Cassie successfully performs the entire motion after using the optimized motion which is the solid line in Fig.~\ref{subfig:stand_opt_loc},\ref{subfig:stand_opt_rot}. Although the elapsed time of the optimized motion is enlarged 2 times compared to the animated one, such change of time span smooths the motion while staying close to the animated motion profile. Such motion profile encodes expressive contents in the animation while also being dynamically feasible. Further, we notice that although there is no motion in the pelvis roll in the animation, the optimization adds a significant movement on that DoF to compensate the fast change of motions from other DoFs. A similar change happens in the lateral transition, though the magnitude is small. This is the reason the optimized motion will not cause failure even though the motion is as aggressive as the animated one. Also note that the entire motion to optimize is 15 seconds and contains 373 nodes. If the animated motion is directly input to the optimization solver without our proposed method, the solver wouldn't be able to find a valid solution. 

\subsection{Experimental Validation}
 \begin{figure*}[t]
    \centering
    \begin{subfigure}{0.37\textwidth}
        \centering
        \includegraphics[width=\linewidth]{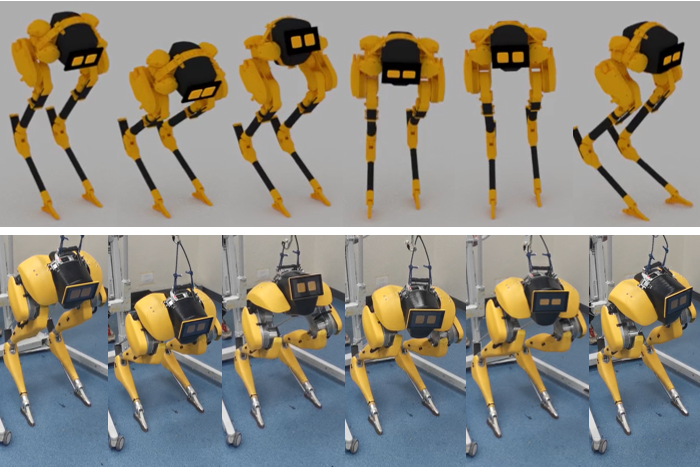}
        \caption{} \label{fig:standing_expt}
    \end{subfigure} 
    \begin{subfigure}{0.29\textwidth}
        \centering
        \includegraphics[width=\linewidth]{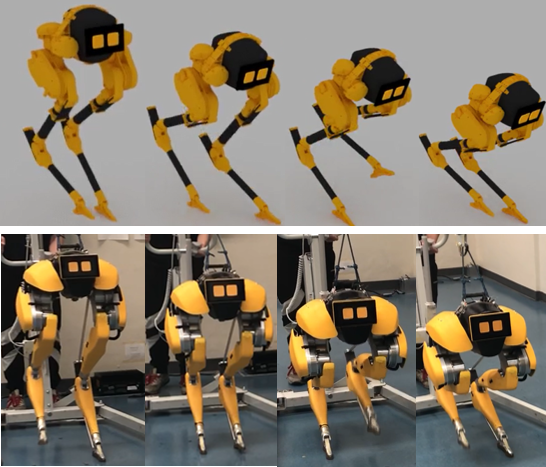}
        \caption{} \label{fig:walking_expt_1}
    \end{subfigure} 
    \begin{subfigure}{0.285\textwidth}
        \centering
        \includegraphics[width=\linewidth]{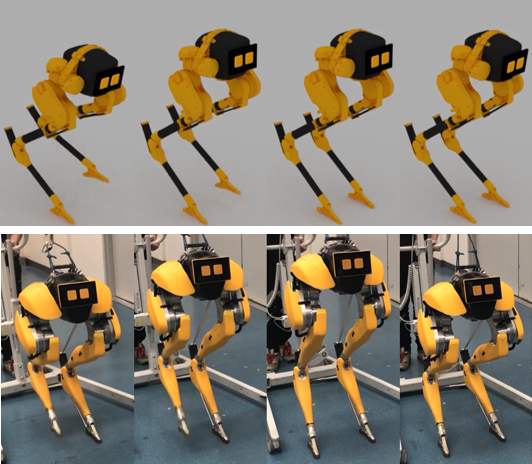}
        \caption{} \label{fig:standing_expt_2}
    \end{subfigure}     
    \caption{Snapshots of Cassie performing curiosity (laser chasing) story. The top half is the original animation and the bottom half is the motions performed by Cassie in real life. (a) Cassie looks on the ground when it is standing. (b) Cassie starts to walk to chase the laser while lowering its walking height. (c) Cassie decelerates and transitions to standing and looks around. The designed eyes videos are also played synchronously with the body motions.}
    \label{fig:experiment}
\end{figure*}

 We deployed and tested the entire proposed pipeline on Cassie. The automaton and gait library interpolation is running at 200Hz and the rest of standing and walking controllers is in a 2kHz loop. 
 We created a story where Cassie shows curiosity by chasing a laser point moving on the ground. This story includes three pieces of aggressive sub-motions, standing-walking-standing, to push the limit of Cassie's mobility. We also created a library containing motions with various emotional attributes, such as happy, sad and tired. We used the laser-chasing story as an illustrative example of the standing-walking-standing motion transition. In this section, we discuss this experiment. The experimental video\footnote{\url{https://youtu.be/zjfzfgb3oK8}} includes the entire story and other emotive motions on Cassie. 
 
 \textit{Standing Section 1}:
 As shown in Fig.~\ref{fig:experiment}a, in the first standing
sequence, Cassie sees a laser point on the ground and leans, turns, and squats to keep it in sight. Such animated movement is aggressive but was optimized to be dynamically feasible.   
 
 \textit{Walking Section}: 
 Cassie then transitions into a walking motion, walking forward while squatting lower while walking to chase the laser. No previous work has been able to adjust Cassie's walking height, as shown in Fig.~\ref{fig:experiment}b.
 
 \textit{Standing Section 2}:
 Finally, Cassie decelerates and stops walking autonomously, transitions back to standing and performs a final standing motion where it looks around, as illustrated in Fig.~\ref{fig:experiment}c. 
 
 According to the experiments, people can clearly infer the robot's emotions and the entire story idea as the motions are designed to be emotive and eye motions are animated and replayed jointly. Moreover, without the proposed automaton, the transitions between standing and walking could cause failure on the robot as there exists discontinuities in robot states and control policies during mode switching. Therefore, although the original animated motion is arbitrarily aggressive, Cassie is able to mimic its virtual character's motion while maintaining balance and retaining the emotive content from the animation using our proposed pipeline.

\section{Conclusions and Future Work} \label{sec:conclusion}
To our knowledge, this paper is one of the first to present a dynamic relatable robotic character to behave with agility, emotions and personalities based on animated motions of a virtual character. In this work, a virtual character that enforces the kinematic constraints of Cassie is created to bridge the animation technology and robotics. A segmented optimization framework is developed to translate arbitrarily long animated motions to be dynamically feasible for the legged robot in both standing and walking scenarios. A versatile walking controller which is able to change the walking height for Cassie is also realized. An end-to-end pipeline centered on a state machine is developed to concatenate different motions from a motion library based on the input commands and feedback from the robot in real time. Experiments are conducted and the animated Cassie shows a capacity to perform motions with emotional attributes. 

In the future, the motion library needs to include more animated and spirited motions enabling human-dynamic-robot interaction based on the emotive robot behaviors created by our proposed animation approach.

\balance
\bibliography{references.bib}
\bibliographystyle{IEEEtran}

\end{document}